\documentclass{article}

\PassOptionsToPackage{numbers, sort&compress}{natbib}

 \usepackage[final]{nips_2018}

\usepackage[utf8]{inputenc} 
\usepackage[T1]{fontenc}    
\usepackage{hyperref}       
\usepackage{url}            
\usepackage{booktabs}       
\usepackage{amsfonts}       
\usepackage{nicefrac}       
\usepackage{microtype}      
\usepackage{graphicx}
\graphicspath{ {./figs/} }
\usepackage{amsmath}
\usepackage{array}
\usepackage{makecell}
\usepackage[table]{xcolor}

\usepackage[linesnumbered,ruled]{algorithm2e}

\title{Unsupervised learning with GLRM feature selection reveals novel traumatic brain injury phenotypes}

\author{
  Aaron J. Masino \\
  Dept. of Anesthesiology \& Critical Care Medicine\\
  Perelman School of Medicine\\
  University of Pennsylvania\\ 
  Philadelphia, PA 19146 \\
  \texttt{masinoa@email.chop.edu} 
   \And
   Kaitlin A. Folweiler \\
   Department of Neurology\\
   Perelman School of Medicine\\
  University of Pennsylvania\\
   Philadelphia PA 19146\\
   \texttt{kfolw@pennmedicine.upenn.edu} 
}

\begin{document}

\maketitle

\begin{abstract}
Baseline injury categorization is important to traumatic brain injury (TBI) research and treatment. Current categorization is dominated by symptom-based scores that insufficiently capture injury heterogeneity. In this work, we apply unsupervised clustering to identify novel TBI phenotypes. Our approach uses a generalized low-rank model (GLRM) model for feature selection in a procedure analogous to wrapper methods. The resulting clusters reveal four novel TBI phenotypes with distinct feature profiles and that correlate to 90-day functional and cognitive status.
\end{abstract}

\section{Introduction}
Traumatic brain injury (TBI) is a leading cause of death and disability in the United States, with an estimated 2.8 million new cases annually \citep{Taylor2017}. In the United States alone, more than 5.2 million individuals currently live with TBI-related disabilities spanning a diverse array of neurobehavioral symptoms such as cognitive, emotional, and motor impairments \citep{Thurman1999}. 

Significant heterogeneity exists within the spectrum of TBI relative to cause, severity, pathology, and prognosis. This heterogeneity represents a major challenge to phenotype categorization which is important to TBI research directed at understanding the underlying physiological mechanisms of injury and recovery, and therapeutic development. Currently, TBI is predominantly categorized by symptom-based scores such as the Glasgow Coma Scale (GCS), which is the primary selection criteria for inclusion in most TBI clinical trials \citep{Saatman2008}. However, there is a consensus among TBI researchers that GCS derived categories are insufficient to capture the complexity of brain injury and do not adequately correlate with long term outcomes \citep{Saatman2008, Hawryluk2016}. 

Unsupervised machine learning is a promising method for data-driven phenotype discovery that may improve current TBI categorization. Unsupervised methods have previously been used to identify patient sub-populations for several diseases \citep{Alizadeh2000, Gamberger2016, Li2015} and have recently been applied within the limited context of mild TBI \citep{Nielson2017, Si2018}. In the work presented here, we sought to extend the use of unsupervised learning in TBI to identify sub-populations with homogeneous clinical characteristics for a broad range of injury severity. Our approach, described in detail below, uses a two-stage pipeline. In the first stage, we perform automated feature selection via generalized low-rank model (GLRM) analysis in a manner analogous to wrapper methods used in supervised learning contexts. In the second stage, the selected feature set is used to cluster the entire dataset.   

\section{Methods}
\subsection{Study cohort data}
Analysis was conducted on data from the Citicoline Brain Injury Treatment Trial (COBRIT; NCT00545662), a clinical trial of 1,213 TBI patients \citep{Zafonte2009, Zafonte2012} \footnote{data available through \href{https://fitbir.nih.gov/}{https://fitbir.nih.gov/}}. Patients were between the ages of 18 and 70 years old, had been diagnosed with a non-penetrating TBI, and had a positive baseline CT scan. Baseline data was acquired prior to injury or within 24 hours after injury and included CT scans, physiological measurements, vital signs, medical history, demographics, laboratory test results, drug screening, and injury information. Where reported, multiple physiological measurements were averaged into a single value. All categorical features were consolidated into a maximum of three categories based on category frequency and dummy coded. Patients with missing values for any baseline feature were excluded, leaving 991 for this study. Outcome data included 12 assessments of 90 day functional and cognitive status. Only patients with complete outcome data were used for analysis of cluster to outcome association (n = 771 at 90 days).

\subsection{Generalized Low-Rank Model feature selection}\label{sec:glrm}
\begin{figure}[ht] 
  \centering
  \fbox{\rule[1cm]{0.5cm}{0cm} 
  \includegraphics[width=0.5\linewidth]{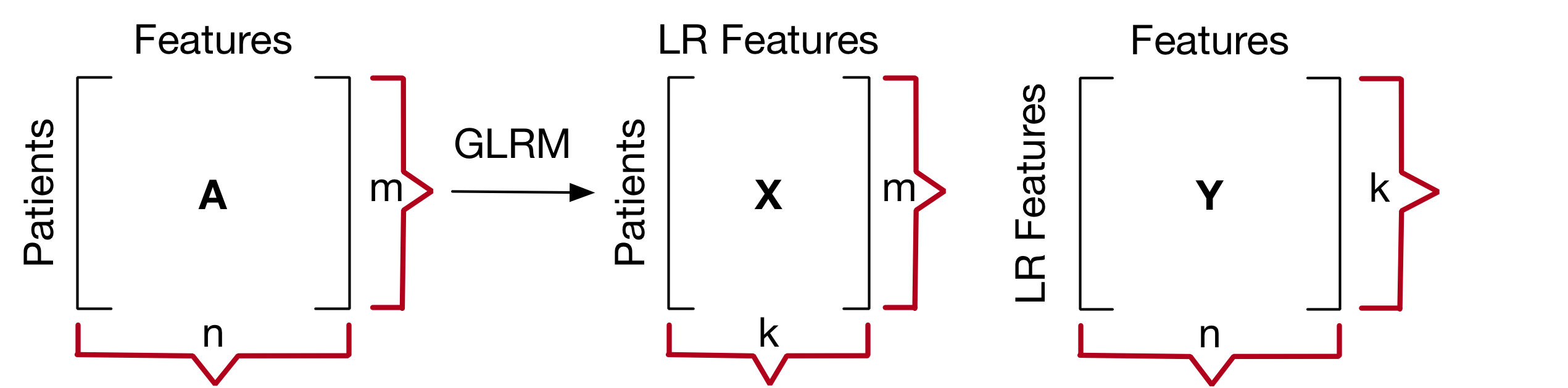}
  \rule[-.1cm]{-.1cm}{-.1cm}}
  \caption{A GLRM model decomposes the original matrix, $\mathbf{A}$, into matrices $\mathbf{X}$ and $\mathbf{Y}$. Assuming $\mathbf{A}$ is a patient data matrix with rows representing patients and columns representing clinical features, the rows of $\mathbf{Y}$ can be viewed as a set of new low-rank (LR) features composed of combinations of the original features and the rows of $\mathbf{X}$ as patient representations in the low-rank feature space.}
  \label{fig:glrm}
\end{figure}

A GLRM decomposes an $m \times n$ matrix, $\mathbf{A}$, into matrices $\mathbf{X}$ and $\mathbf{Y}$ such that $\mathbf{X}\mathbf{Y}$ is approximately equal to $\mathbf{A}$ under the constraint that the number, $k$, of linearly independent columns in $\mathbf{X}$ (i.e. the rank) satisfies $k<n$ (see Figure \ref{fig:glrm}). Importantly for biomedical applications, GLRM's can represent high-dimensional data of mixed data types in a transformed lower-dimensional space \citep{Udell2016}. We used a GLRM to decompose the patient data matrix, $\mathbf{A}$, composed of $m$ patient rows and $n$ clinical feature columns.  We fixed our low-rank model at rank $k=2$, and used quadratic and hinge loss functions to approximate numerical and binary features, respectively \citep{Udell2016}. We applied L1-norm regularization on $\mathbf{Y}$ to reduce the size of the feature set contributing to the low rank decomposition, resulting in the following loss function:
\begin{equation}
    L\left(\mathbf{A},\mathbf{XY}\right)=
    \left\{ \begin{array}{l l}
      \sum\limits_{i=1}^{m} \sum\limits_{j=1}^{n} \nicefrac{1}{\sigma_j^2} \left(\mathbf{A}_{ij}-\mathbf{X}_i\mathbf{Y}_j\right)^2 + \gamma \sum\limits_{j=1}^n  \left|\left| \mathbf{Y}_j\right|\right|_1
      &  \mbox{if } \mathbf{A}_{:,j} \in \mathbb{R}\\
      \sum\limits_{i=1}^{m} \sum\limits_{j=1}^{n} \max
      \left(0, 1-\mathbf{A}_{ij}\cdot\mathbf{X}_i\mathbf{Y}_j\right) + \gamma \sum\limits_{j=1}^n  \left|\left| \mathbf{Y}_j\right|\right|_1
      & \mbox{if } \mathbf{A}_{:,j} \in \left\{0,1\right\} 
      \end{array}\right.  
\end{equation}
where the L1-regularization parameter, $\gamma$, is greater than zero, and $\mathbf{A}_{:,j}$ indicates column $j$ of $\mathbf{A}$. The quadratic loss function was scaled by the inverse variance, $\nicefrac{1}{\sigma_j^2}$, of each feature, $j$, to compensate for unequal scaling in different features.  Regularization produces a column-sparse matrix, $\mathbf{Y}$, where the number of non-zero columns, $d$, is small relative to the total column number, $n$. If $\mathbf{Y}_{:,j}$ is a zero vector, then feature $j$ from $\mathbf{A}$ is not used in the formulation of \emph{any} of the low-rank features which suggests it was relatively uninformative in approximating $\mathbf{A}$ \citep{Witten, Witten2009, Shen2007}. Therefore, for a given choice of $\gamma$, we select the $d$ features from the \emph{original input space} for which there is at least one non-zero entry in the column $\mathbf{Y}_{:,j}$. In this manner, the GLRM is utilized for feature reduction, rather than transformation into a lower dimensional space. 

\subsection{Clustering method}\label{sec:clustering}
Given a feature set, we calculated the pairwise dissimilarity between observations using Gower's measure \citep{Gower1971} to form a dissimilarity matrix. Gower's measure was selected because it accommodates mixed data types. Weights were added to binary features in pairwise Gower’s coefficient calculations such that both binary and numerical features were weighted proportionally on average. The partitioning around medoids (PAM) algorithm was used to cluster the dissimilarity matrix \citep{Kaufman1990}.

\subsection{Cross-validation feature selection procedure}
To avoid over-fitting, we implemented a $K$-fold cross-validation feature selection procedure to tune the GLRM regularization parameter, $\gamma$ (see Algorithm \ref{alg:crossvalidation}). For each fold, an optimal parameter is determined on the trainig folds. Features are then selected using GLRM on the validation fold. The final feature set was taken as the intersection of features selected across the $K$ validation folds.

\begin{algorithm}
    \SetKw{Kwin}{in}
    \SetKwInOut{Input}{Input}

    \Input{data: $D$; initial gamma: $\gamma_0$; gamma step size: $\gamma_{step}$; complete feature list: all\_features}
    final\_features = all\_features\; 
    \For{$k$ in $\left[1, \ldots, K\right]$}{
        validation\_set = fold($k$,$D$); \quad train\_set = $\bigcup^K_{j=1, j\neq k}$ fold($j$, $D$)\;
        
        increment\_$\gamma$ = True; \quad best\_score = -1; \quad $\gamma = \gamma_0$; \quad best\_train = \{\}\;
        \While{increment\_$\gamma$}{
            increment\_$\gamma$ = False; \quad features = GLRM\_features(train\_set, $\gamma$)\;
            \For{$n_c$ in $\left[n_{min}, \ldots, n_{max}\right]$}{
                clusters = PAM(train\_set, features, $n_c$); \quad score = silhouette(clusters)\;
                \If{score > best\_score}{
                    increment\_$\gamma$ = True; \quad best\_score = score\;
                    update\_results(best\_train, $k$, $\gamma$, score, clusters, features); 
                }
            }
            $\gamma$ += $\gamma_{step}$; 
        }
        best\_score = -1; \quad best\_validation = \{\}\;
        features = GLRM\_features(validation\_set, $\gamma$); \quad \# use $\gamma$ from best\_training\;
        \For{$n_c$ in $\left[n_{min}, \ldots, n_{max}\right]$}{
            clusters = PAM(validation\_set, features, $n_c$); \quad score = silhouette(clusters)\;
            \If{score > best\_score}{
                    best\_score = score\;
                    update\_results(best\_validation, $k$, $\gamma$, score, clusters, features)\;
                }
        }
        final\_features = final\_features $\bigcap$ best\_validation['features', 'k'] \# best features fold($k$)
    } 
        \caption{Cross-validation feature selection. \textbf{Functions}: update\_results - stores best values in dictionary; GLRM\_features - feature selection per section \ref{sec:glrm}; PAM - clustering per section \ref{sec:clustering}}  
    \label{alg:crossvalidation}
\end{algorithm}

\subsection{Cluster analysis}
Training and validation cluster stability was assessed by comparing the similarity of cluster membership between observation pairs in the training and validation folds, respectively, with their cluster membership in the full dataset using the Pairwise Similarity Index. The Pairwise Similarity Index for two equal length sets of clustering labels, Set A and Set B, is the percent of total observation pairs that belong to the same cluster in both Set A and Set B.

\subsection{Statistical Analysis}
To compare differences in feature values between clusters, group-wise comparisons were assessed using the Kruskal-Wallis test for non-parametric one-way analysis of variance. When the Kruskal-Wallis test indicated overall significance, the Holm multiple comparison test was used to determine specific group differences where applicable. Pearson’s chi-squared test was used to compare distributions of categorical variables. Statistical comparisons were considered significant when p values were < 0.05 or as determined by the Holm’s test statistic for multiple comparisons. 

\section{Results}
From the 98 features available in the COBRIT dataset, our feature selection process retained 13: net fluid intake; heart rate; blood count measures of hematocrit, hemoglobin, platelets, white blood cells; blood levels of glucose and the liver enzyme aspartate aminotransferase (AST); and clotting indicators prothrombin time (PT), partial thromboplastin time (PTT) and PT International Normalized Ratio (INR); presence of abnormal mesencephalic cisterns on the CT scan (whether blood-filled, compressed or obliterated); indicator of falling as injury mechanism. The \emph{necessity} of each feature was evaluated by randomly shuffling its values between patients, clustering the data and comparing the results to the clusters obtained using the non-shuffled values via the Jaccard similarity and pairwise similarity index. This process was repeated 500 times. Only INR was found to be unnecessary and was removed for the final analysis. 

\begin{figure}[ht] 
  \centering
  \fbox{\rule[1cm]{0.5cm}{0cm} 
  \includegraphics[height=0.3\linewidth]{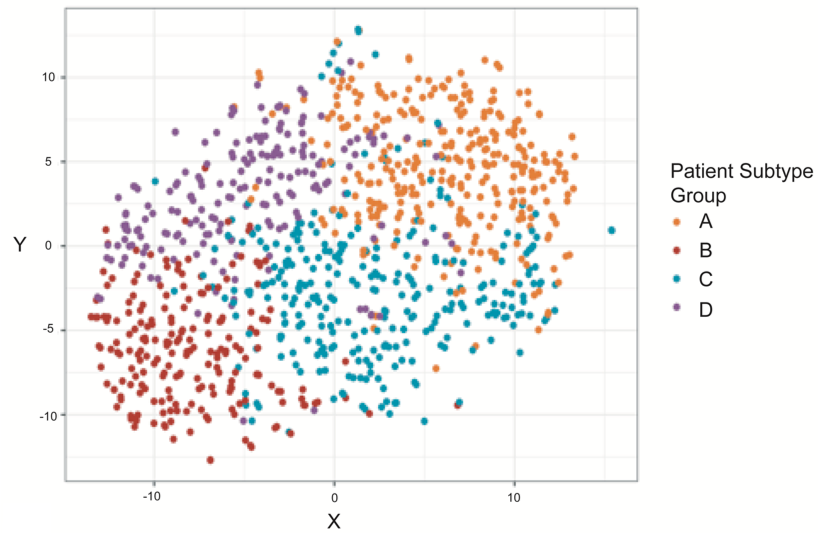}
  \rule[-.1cm]{-.1cm}{-.1cm}}
  \caption{T-SNE projection of clusters found with PAM clustering and GLRM feature selection. Cluster patient membership counts: (A, 314), (B, 204), (C, 286), and (D,187).}
  \label{fig:clusters}
\end{figure}

The full dataset of 991 patients was clustered using the remaining 12 features resulting in four distinct patient phenotypes, as shown in Figure \ref{fig:clusters}. Statistically significant differences between clusters were found for all 12 baseline features (Appendix Table \ref{table:baselineProfiles}). At 90 days, there were significant differences between clusters on 9 out of 12 of the outcome assessment scores (Appendix, Table \ref{table:outcome90Profiles}). For comparison, when patients were grouped by their GCS score at baseline, the current clinical standard, no significant baseline feature or 90 day outcome differences were detected between groups.

\section{Discussion \& conclusion}
We identified 4 TBI patient phenotypes using unsupervised machine learning. Each phenotype possess a unique baseline feature profile which corresponded to differences in long-term outcomes. Comparatively, when categorized by GCS score, distinct feature profiles are not observed and categories did not correlate with long-term outcomes. Methodologically, we used a GLRM to identify important features from the original input space. This is advantageous for biomedical applications because it enables feature reduction to control over-fitting, while retaining interpretable clinical features. The selected features plausibly influence TBI outcome. Platelet and hemoglobin counts, hematocrit, PT, PTT, \& net fluid intake contribute to coagulation which is common in TBI and is associated with poor outcomes  \citep{Harhangi2008,Hulka1996,Stein2004}. High net fluid intake, low complete blood count measures and prolonged PT/PTT duration are indicative of possible hemorrhage \citep{Talving2009}, which may have occurred in patients with abnormal mesencephalic cisterns which were more represented in subtype B. Abnormal cistern imaging, elevated AST levels, and increased white blood cell counts have been associated with TBI in previous research \citep{Nielson2017,Rovlias2001,Hartl1997}. Falling as injury cause, the only selected non-physiological feature, is associated with intracranial pressures that differ from other injury mechanisms \citep{Post,Yanagida}.

Study limitations include: all patients had positive CT findings, which is not true of all TBIs; baseline features did not include genetic data or other neuroimaging modalities; phenotype profile and patient outcome realations are only correlative. These areas are considerations for our future research. 

Our results demonstrate that unsupervised machine learning may advance TBI translational research. With refinement, we anticipate that data-derived patient phenotypes will supplement existing clinical assessments for TBI patient classification in clinical trials and provide clinical decision support. 



\bibliographystyle{unsrtnat}
\bibliography{main}

\newpage
\section*{Appendix}\label{appendix}

\definecolor{subtlegrey}{rgb}{0.9, 0.95, 0.91}
\definecolor{white}{rgb}{1.0, 1.0, 1.0}

\begin{table}[h!]
\small
  \caption{Injury baseline feature profile comparison between TBI phenotypes discovered by unsupervised learning with GLRM feature selection. Kruskal-Wallis Test p-value < 0.0001 for all continuous valued features. Pearson's Chi Squared Test p-value < 0.001 for all binary features.}
  \label{table:baselineProfiles}
  \centering
 \rowcolors{3}{subtlegrey}{white}
 
 \begin{tabular}{m{10em} c c c c c c c c}
    \toprule
    & \multicolumn{2}{c}{\textbf{Subtype A}} & \multicolumn{2}{c}{\textbf{Subtype B}} &\multicolumn{2}{c}{\textbf{Subtype C}} & \multicolumn{2}{c}{\textbf{Subtype D}} \\
    \bottomrule
    
    \thead[bc]{\textbf{Clinical Features}\\\textbf{Continuous}} & \textbf{Median} & \textbf{IQR} & \textbf{Median} & \textbf{IQR} & \textbf{Median} & \textbf{IQR} & \textbf{Median} & \textbf{IQR} \\
    \bottomrule
    \makecell[bl]{Platelet count (1000/$\mu L$)} & 219.0 & 80.0 & 181.5 & 91.3 & 216.0 & 80.0 & 177.4 & 71.1 \\
    \makecell[bl]{Hemoglobin (g/dL)} & 14.1 & 1.8 & 10.3 & 2.2 & 13.0 & 1.9 & 11.1 & 1.9 \\ 
    \makecell[bl]{AST (IU/L)} & 33.0 & 19.0 & 83.5 & 96.3 & 50.0 & 60.5 & 40.0 & 28.8 \\
    \makecell[bl]{Net Fluid Intake (L)} & 1.7 & 1.5 & 5.7 & 4.1 & 3.1 & 2.2 & 3.4 & 2.4 \\
    \makecell[bl]{Heart Rate (bpm)} & 81.0 & 14.3 & 103.5 & 17.8 & 97.5 & 15.5 & 79.5 & 12.9 \\
    \makecell[bl]{Glucose (mg/dL)} & 120.5 & 29.8 & 152.8 & 45.1 & 136.0 & 35.4 & 134.8 & 38.3 \\
    \makecell[bl]{White Blood Cell Count\\(1000/$\mu L$)} & 11.6 & 5.4 & 16.5 & 6.9 & 14.3 & 5.2 & 12.5 & 5.5 \\
    \makecell[bl]{Hematocrit (\%)} & 41.0 & 4.5 & 32.5 & 5.5 & 39.0 & 5.0 & 35.3 & 4.9 \\
    \makecell[bl]{Prothrombin Time (sec)} & 13.0 & 2.9 & 14.8 & 3.1 & 10.9 & 2.9 & 13.8 & 3.4 \\
    \makecell[bl]{Partial Thromboplastin\\Time (sec)} & 26.5 & 4.3 & 28.5 & 5.5 & 25.0 & 3.9 & 27.0 & 4.4 \\
    \bottomrule
    \bottomrule
     \rowcolor{white}
    \thead[bc]{\textbf{Clinical Features}\\\textbf{Binary}} & \multicolumn{2}{c}{\textbf{\% Count}} & 
    \multicolumn{2}{c}{\textbf{\% Count}} &\multicolumn{2}{c}{\textbf{\% Count}} & \multicolumn{2}{c}{\textbf{\% Count}} \\
    \bottomrule
     \rowcolor{subtlegrey}
    \makecell[bl]{Abnormal Mesencephalic\\Cisterns (CT Finding)} & \multicolumn{2}{c}{15.9\% (50)}  & \multicolumn{2}{c}{44.1\% (90)}  & \multicolumn{2}{c}{25.2\% (72)}  & \multicolumn{2}{c}{63.1\% (118)} \\
     \rowcolor{white}
    \makecell[bl]{Fall Injury Mechanism} &  \multicolumn{2}{c}{28.3\% (89)}  & \multicolumn{2}{c}{90.0\% (184)}  & \multicolumn{2}{c}{78.3\% (224)}  & \multicolumn{2}{c}{35.2\% (66)}\\
    \bottomrule
  \end{tabular}
\end{table}

\begin{table}[h!]
\small
  \caption{90-day functional and cognitive status comparison between injury baseline TBI phenotypes discovered by unsupervised learning with GLRM feature selection. Kruskal-Wallis Test p-value indicated in last column.}
  \label{table:outcome90Profiles}
  \centering
  \rowcolors{3}{subtlegrey}{white}
 \begin{tabular}{m{10em} c c c c c c c c c}
    \toprule
    & \multicolumn{2}{c}{\textbf{Subtype A}} & \multicolumn{2}{c}{\textbf{Subtype B}} &\multicolumn{2}{c}{\textbf{Subtype C}} & \multicolumn{2}{c}{\textbf{Subtype D}} \\
    \bottomrule
    
    \thead[bc]{\textbf{Primary Outcomes}} & \textbf{Median} & \textbf{IQR} & \textbf{Median} & \textbf{IQR} & \textbf{Median} & \textbf{IQR} & \textbf{Median} & \textbf{IQR} & \textbf{p-value} \\
    \bottomrule
    \makecell[bl]{Glasgow Outcome Scale\\Extended} & 6 & 1 & 4 & 3 & 5 & 2 & 4 & 3 & <0.0001 \\
    \makecell[bl]{California Verbal\\Learning Test} & 49 & 17 & 39 & 21 & 47 & 20 & 42 & 20 & <0.0001 \\ 
    \makecell[bl]{Processing Speed Index} & 26 & 28 & 24 & 30 & 29 & 27 & 23.5 & 30.3 & 0.0044 \\
    \makecell[bl]{Trial Making A} & 43 & 18 & 29 & 40 & 39 & 21 & 37 & 27.3 & <0.0001\\
    \makecell[bl]{Trail Making B} & 42 & 19 & 33 & 40 & 39 & 28 & 37.5 & 26.5 & <0.0001\\
    \makecell[bl]{Digit Span} & 9 & 3 & 8 & 3 & 9 & 4 & 9 & 3.8 & 0.0328\\
    \makecell[bl]{Stroop Part 1} & 45.5 & 15.3 & 51 & 18 & 49 & 17 & 48 & 16 & 0.0005\\
    \makecell[bl]{Stroop Part 2} & 112 & 42 & 112 & 63 & 110 & 52 & 106.5 & 52.5 & 0.5319\\
    \makecell[bl]{Controlled Oral Word\\Association Test} & 38 & 14 & 31 & 15 & 34 & 15 & 34.5 & 14.3 & <0.0001\\
    \bottomrule
    \bottomrule
    \thead[bc]{\textbf{Secondary Outcomes}} & \textbf{Median} & \textbf{IQR} & \textbf{Median} & \textbf{IQR} & \textbf{Median} & \textbf{IQR} & \textbf{Median} & \textbf{IQR} & \textbf{p-value} \\
    \bottomrule
    \makecell[bl]{Disability Rating Scale} & 0 & 1 & 4 & 6 & 1 & 3 & 1 & 4 & <0.0001 \\
    \makecell[bl]{Satisfaction with Life\\Scale} & 24 & 12 & 21 & 10.8 & 23 & 11 & 23 & 11 & 0.6702\\
    \makecell[bl]{Brief Symptoms\\Inventory-18 Global\\Severity Index} & 56 & 19.8 & 57 & 14.8 & 55 & 18 & 56 & 15.5 & 0.7566\\
    \bottomrule
  \end{tabular}
\end{table}

\end{document}